\documentclass[10pt,twocolumn,letterpaper]{article}

\usepackage[final,applications]{wacv} 
\usepackage{times}
\usepackage{epsfig}
\usepackage{graphicx}
\usepackage{amsmath}
\usepackage{amssymb}
\usepackage{booktabs}
\usepackage{adjustbox}
\usepackage{multirow}
\usepackage[table,xcdraw,dvipsnames]{xcolor}
\usepackage{color, colortbl}
\usepackage{nicefrac}
\usepackage[accsupp]{axessibility}

\definecolor{Gray}{gray}{0.9}
\definecolor{brightturquoise}{rgb}{0.85, 1, 1}
\definecolor{newpurple}{HTML}{BC61F5}

\newcommand{\reffig}[1]{\text{Figure~\ref{#1}}}
\newcommand{\reftab}[1]{\text{Table~\ref{#1}}}
\newcommand{\refeq}[1]{\text{Eq.~\ref{#1}}}
\newcommand*{\affaddr}[1]{#1}
\newcommand*{\affmark}[1][*]{\textsuperscript{#1}}

\newcommand\blfootnote[1]{%
  \begingroup
  \renewcommand\thefootnote{}\footnote{#1}%
  \addtocounter{footnote}{-1}%
  \endgroup
}

\usepackage[pagebackref=true,breaklinks=true,colorlinks,bookmarks=false,citecolor=blue,linkcolor=red]{hyperref}

\begin{document}

\title{Learning Saliency From Fixations}

\author{Yasser Abdelaziz Dahou Djilali\affmark[1,2] \quad 
Kevin McGuiness \affmark[1] \quad 
Noel O'Connor\affmark[1] \vspace{0.1cm}\\
\affaddr{\affmark[1]Dublin City University, Ireland} \quad 
\affaddr{\affmark[2]Technology Innovation Institute, UAE}
}

\maketitle
\thispagestyle{empty}

\begin{abstract}

We present a novel approach for saliency prediction in images, leveraging parallel decoding in transformers to learn saliency solely from fixation maps. Models typically rely on continuous saliency maps, to overcome the difficulty of optimizing for the discrete fixation map. We attempt to replicate the experimental setup that generates saliency datasets. Our approach treats saliency prediction as a direct set prediction problem, via a  global loss that enforces unique fixations prediction through bipartite matching and a transformer encoder-decoder architecture. By utilizing a fixed set of learned fixation queries, the cross-attention reasons over the image features to directly output the fixation points, distinguishing it from other modern saliency predictors. Our approach, named Saliency TRansformer (SalTR), achieves metric scores on par with state-of-the-art approaches on the Salicon and MIT300 benchmarks.

\blfootnote{Code: \url{https://github.com/YasserdahouML/SalTR}}

\end{abstract}

\section{Introduction}

 In recent years, deep learning models have achieved significant progress in saliency prediction ~\cite{pan2017salgan, borji2019saliency}, leveraging large-scale annotated datasets \cite{mit-saliency-benchmark, jiang2015Salicon, borji2015cat2000}. A saliency dataset collection requires a set of images serving as visual stimuli, along with recorded eye movements data captured through eye-tracking devices. Generally, each stimuli is observed by several human subjects. During the observation, the eye positions of the subjects are continuously tracked in relation to the coordinates of the images to obtain the fixation maps. Then, the individual fixations are aggregated and blurred with a Gaussian filter to generate a continuous saliency map \cite{xu2014predicting}.
 
 Most existing saliency prediction models rely on continuous saliency maps, where each pixel represents the probability of attending at that location in the image \cite{pan2016shallow, kruthiventi2017deepfix, pan2017salgan}. However, these models face challenges in optimizing for the discrete fixation maps, which indicate the specific locations of saccades by human observers. While successful in producing high quality saliency maps, the common architecture does not replicate the data collection pipeline, but rather tries to implicitly learn through the post-processed saliency map. To overcome this limitation, and inspired from \cite{carion2020end}, we propose a novel approach that learns saliency prediction solely from fixation maps, without relying on continuous saliency annotations. Our method, named Saliency TRansformer (SalTR), leverages parallel decoding in transformers to directly predict the fixation points directly.

In our approach, we treat saliency prediction as a direct set prediction problem, where the goal is to predict a set of spatial fixation points. To achieve this, we employ a transformer encoder-decoder architecture \cite{vaswani2017attention}, with a fixed set of learned fixation queries. The cross-attention mechanism in the transformer decoder reasons over the image features using these fixation queries to directly output the fixation points. This distinguishes our approach from other modern saliency predictors, which typically rely on continuous saliency maps. In summary, our contributions are: 
\begin{itemize}
    \item We propose a novel approach for saliency prediction, leveraging parallel decoding in transformers to learn saliency solely from fixation maps.

    \item We demonstrate the effectiveness of our approach on the Salicon benchmark, achieving remarkable performance compared to state-of-the-art methods.

    \item Furthermore, we extend the approach to the scanpaths prediction problem, and demonstrates its effectiveness.
\end{itemize}

\section{Related works}

  The seminal work of the feature integration theory \cite{treisman1980feature}, is a cornerstone in identifying the visual features that guide human attention. This foundational theory has served as a launchpad for the development of various computational models. In the realm of computer vision, the emphasis is primarily placed on the selective mechanism when modeling attention. Saliency, in this context, is subtly defined in relation to the gaze policy on a scene – characterizing the particular subsets of space where a human observer would likely concentrate their focus. The term "salient" surfaced in the sphere of bottom-up computations \cite{koch1987shifts,itti1998model}, while the concept of attention spans a broader spectrum. Furthermore, the last decade has witnessed the remarkable progress of saliency prediction, and many methods have been presented and achieved remarkable performances on the recently introduced benchmarks, especially the deep learning based methods have yielded a boost in performance. Researchers tend to typically repurpose existing Convolutional Neural Networks (CNN) architectures \cite{o2015introduction, simonyan2014very, he2016deep} to make predictions about saliency. These models involve architectural enhancements tailored to the specific demands of the saliency downstream task. These models are trained end-to-end on saliency datasets, framing saliency as a regression problem. A common challenge faced in this area is the scarcity of annotated fixation data. To mitigate this issue, the model's encoder is usually pretrained on extensive image recognition datasets, such as ImageNet \cite{russakovsky2015imagenet}. This pretraining step allows for the acquisition of valuable representations at the level of latent space, which are then fine-tuned on saliency datasets. Prior to that, handcrafted approaches attempted in modelling the human visual attention.

\textbf{Heuristic approaches.} The prediction of saliency for images has been a major focus of academic research over the past few decades. A seminal study by ~\cite{itti1998model} introduced a bottom-up approach to visual saliency, utilizing center-surround differences across multiple scales of image features. This method generates conspicuity maps by linearly combining and normalizing feature maps, with color (C), orientation (O), and intensity (I) serving as the three primary features:

\begin{equation}
C_{I}=f_{I},\quad C_{C}={\cal{N}}(\sum_{l\in L_{C}}f_{l}),\quad C_{O}={\cal{N}}(\sum_{l\in L_{O}}f_{l}).
\end{equation}

In this equation, ${\cal{N}}(.)$ represents the map normalization operator. The ultimate saliency map is an average of the three conspicuity maps: $S=\frac{1}{3}\sum_{k \in {I,C,O}}C_{k}$. A more complex bottom-up saliency model, taking into account additional Human Visual System (HVS) features such as contrast sensitivity functions, perceptual decomposition, visual masking, and center-surround interactions, was later proposed in ~\cite{le2006coherent}. Additional static saliency models, such as those developed by \cite{navalpakkam2006integrated,kootstra2008paying,murray2011saliency,bruce2006saliency,garcia2009decorrelation,seo2009static,goferman2012context,gao2005discriminant}, are predominantly cognitive-based models. These models utilize various visual features, including color, edge, and orientation, at numerous spatial scales to construct a saliency map.

In addition, Bayesian models have been employed to supplement these cognitive models, introducing a layer of prior knowledge (e.g., scene context or gist) through a probabilistic approach such as Bayes' rule for combination ~\cite{torralba2003modeling,oliva2003top,zhang2008sun,ehinger2009modelling}. These models demonstrate the capacity to integrate various factors in a principled manner.

\textbf{Deep learning approaches.} are rooted in data-driven approached from recorded eye-fixations or labeled salient maps. Authors from \cite{kienzle2009center} pioneered a non-parametric bottom-up method to learn saliency from human eye fixation data. Their model utilizes a support vector machine (SVM)  \cite{hearst1998support} to determine saliency based on local intensities, marking the first method that did not rely on any assumptions about Human Visual System (HVS) features to encode saliency. Similarly, Judd et al. \cite{judd2009learning} used a linear SVM to train on 1003 labeled images, leveraging a range of low, mid, and high-level image features.

Recent deep learning-based static saliency models, such as those proposed by ~\cite{ kummerer2014deep, 10.1007/978-3-030-68796-0_22,kruthiventi2017deepfix,cornia2016deep,pan2016shallow,pan2017salgan}, have made notable advancements, leveraging the success of deep neural networks and the availability of large-scale saliency datasets for static scenes, such as those described in ~\cite{mit-saliency-benchmark,borji2015cat2000}.

The works of \cite{pan2016shallow, kruthiventi2017deepfix} were pioneers in the application of Convolutional Neural Networks (CNNs) for saliency prediction, culminating in the creation of the eDN and DeepFix models respectively. Specifically, DeepFix employs a unique approach in its initial phase, utilizing the weights from the first five convolution blocks of the VGG-16 model \cite{simonyan2014very}. Furthermore, it introduces two Location Based Convolutional (LBC) layers, adept at capturing semantics at various scales. Subsequently, Pan et al.~\cite{pan2017salgan} leveraged Generative Adversarial Networks (GANs) ~\cite{goodfellow2014generative} to devise the SalGAN model. The SalGAN architecture comprises a generator model, the weights of which are learned through back-propagation. This learning is driven by a binary cross-entropy (BCE) loss computed over pre-existing saliency maps. The ensuing prediction generated by the model is further processed by a discriminator network.

This discriminator network is trained to perform a binary classification task, tasked with distinguishing between the saliency maps generated by the generator and the ground truth maps. The process follows a min-max game format, utilizing the ensuing adversarial loss:

\begin{equation}
{\cal{L}}= \alpha{\cal{L}}_{\text{BCE}}+ \mathcal{L}(D(q,p_{model}),1).
\end{equation}

In this equation, the aim is to optimize the value of $\mathcal{L}(D(I,S),1)$. Here, $D(I, S )$ represents the probability that the discriminator network is deceived, meaning that the generated saliency maps closely mimic those from the data distribution ground truth. Authors from ~\cite{bylinskii2016should} proposed SALICON, the model optimizes an objective function based on the saliency evaluation metrics, from two parallel streams at different image scales. \cite{liu2018deep} trained the deep spatial contextual long-term recurrent convolutional network (DSCLRCN), incorporating both global spatial interconnections and scene context modulation. EML-NET proposed by \cite{jia2020eml} consists of a disjoint encoder and decoder trained separately. Furthermore, the encoder can contain many networks extracting features, while the decoder learns to combine many latent variables generated by the encoder networks. Unisal \cite{droste2020unified}  introduced four domain adaptation techniques aimed at addressing the significant challenge of disparity in datasets for effective joint modeling. These strategies encompass Domain-Adaptive Priors, Domain-Adaptive Fusion, Domain-Adaptive Smoothing, and Bypass-RNN. The model also proposes a refined formulation of learned Gaussian priors. These techniques were incorporated into a streamlined, lightweight network constructed in the encoder-RNN-decoder style. SalFBNet \cite{ding2022salfbnet}, builds on the benefits of feedback connections by linking higher-level feature blocks to low-level layers. The study also proposes a novel Selective Fixation and Non-Fixation Error loss, a large-scale Pseudo-Saliency dataset, and shows that SalFBNet performs competitively on public benchmarks with fewer parameters, indicating the effectiveness of their approach. Through principled combination of multiple backbones, they developed a the current SoTA model, "DeepGaze IIE \cite{linardos2021deepgaze}", that achieves the best performance on the MIT/Tuebingen Saliency \cite{kummerer2018saliency} Benchmark across all metrics, highlighting the model's ability to maintain good confidence calibration on unseen datasets. Overall, these deep models achieve results closer to the human baseline results on the SALICON~\cite{bylinskii2016should}, MIT300~\cite{bylinskii2015saliency}, and CAT2000~\cite{borji2015cat2000} datasets.

\textbf{DEtection TRansformers.}  DETR \cite{carion2020end} shifted the object detection paradigm, casting the problem as a set-based prediction one, eliminating the hand-designed components, and maintaining comparable performance against the well-established FasterRCNN \cite{ren2015faster}. DETR \cite{carion2020end} combines several techniques such as bipartite matching loss, transformer encoder-decoder with parallel decoding to design DEtection Transformer (DETR). The approach formulates the object detection task as an image-to-set problem. The model outputs a fixed-length unordered set of classes and bounding boxes of all possible objects present in the image. The bipartite matching forces unique one-to-one predictions. Intuitively, the decoder queries can be interpreted as humans saccading at various spatial locations of an image; each human hence observes others before making its prediction.

It can be seen from the above review that saliency prediction models mostly use the continious saliency maps for learning. Inspired by DETR, we attempt to learn saliency from fixations only.

\section{Method}
\begin{figure*}
\makebox[\linewidth]{
    \centering
    \includegraphics[width=1.00\linewidth]{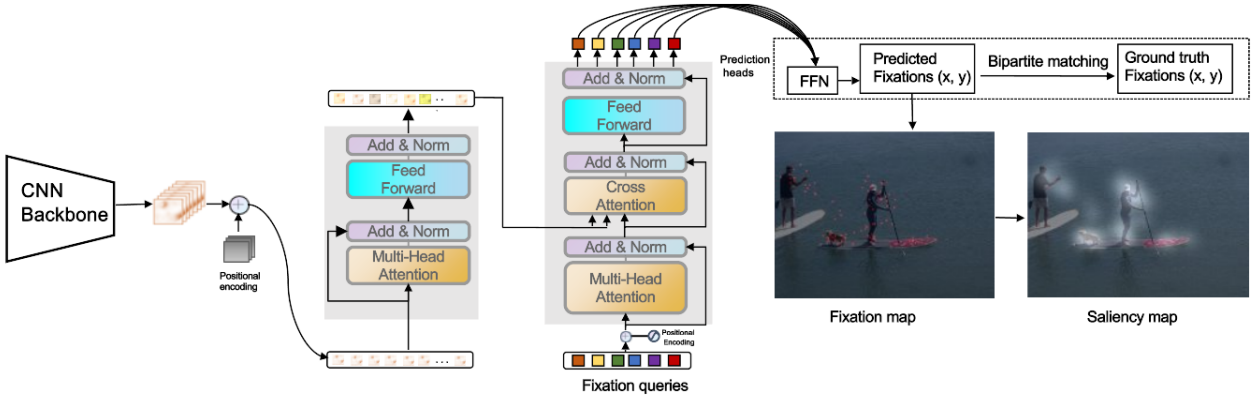}}
    \caption{Complete pipeline for training. The CNN backbone produces the latent representation given an input image. The transformer encoder enhances this representation for a suitable decoding. The queries in the transformer decoder cross-attention are a fixed number of fixation queries, that attend to the image features.  The prediction head maps the output embedding to a spatial fixation location.}
    \label{main_fig}
\end{figure*}

Following DETR \cite{carion2020end}, our model is an end-to-end saliency predictor which includes a ResNet backbone \cite{he2016deep}, Transformer encoder and decoder \cite{vaswani2017attention}, and a fixations prediction head. We  adapt the decoder part to solve the saliency task, as shown in \reffig{main_fig}. Given an image, we extract the latent representations using a CNN backbone followed with a Transformer encoder to enrich the CNN features. Then, the fixation queries are fed to the Transformer decoder to search for fixation locations given the image information through cross attention.

\subsection{SalTR components}

Given an image dataset, $\mathcal{D} = \{ \mathbf{x}_{1},....,\mathbf{x}_{|\mathcal{D}|}\}$ where $\mathbf{x}_{i} \in \mathcal{R}^{C \times H \times W}$, the goal is to predict a set of spatial fixation points $\bar f_{i} = (\bar x_{i},\bar y_{i})$, that will serve as the basis to build the fixation map $\mathcal{M}_{p}$. We then smooth this map with a Gaussian filter with a standard deviation $\sigma$ to obtain the final continious saliency map. However, there are some significant differences that distinguish our approach from previous works. As the model predicts the fixation points only, we do not leverage the continious ground truth saliency map at training time. Thus, we aim at mimicking the way visual attention datasets are created, with the use of both the fixation queries and parallel decoding. The key modules are outlined in the following.

\textbf{CNN Encoder} ($\mathbf{f}_{\theta}$). The encoder is a network $\mathbf{f_{\theta}}: \mathbf{x} \mapsto \boldsymbol {\Gamma}$ parameterised by $\boldsymbol{\theta_{e}}$.  $\mathbf{f}_{\theta}$ is implemented as a backbone ResNet50 \cite{he2016deep}, followed by a 2-layer $1 \times 1$ convolutional projection head with batch normalization and ReLU activation, that reduces the channel dimension from 2048 to 256.

\textbf{Transformer encoder}.  $\boldsymbol{\Omega_{\omega}}$ maps  $\boldsymbol {\Gamma}^{c \times h \times w}$ to $\boldsymbol {\gamma}^{c \times h \times w}$.   $\boldsymbol {\Gamma}$ is first wrapped to a sequence of size $c \times hw$, then augmented with 2D positional encodings \cite{bello2019attention}. The multi-head self-attention layers perform message parsing across $\boldsymbol {\Gamma}$ channels, in order to capture the contextual information. This acts as a smoothing prior, hence, pixels sharing the same semantic class repulsively attend irrespective of their position in the image, ignoring all structural information. Furthermore, $\boldsymbol {\gamma}$ is the smoothed transform of  $\boldsymbol {\Gamma}$, ensuring coherence both spatially in the neighborhood of a given pixel $i$ and semantically for pixels $j$ further away but sharing the same class. 

\textbf{Transformer decoder}. The decoder transforms a fixed number of embeddings (i.e.\ fixation queries) of size $256$ using multi-headed cross attention mechanisms, where the keys and values are sourced from the image features. The output embeddings are then mapped to a fixation point $\bar f_{i} = (\bar x_{i},\bar y_{i})$ using a 3-layer MLP.

\subsection{Learning from fixations}
Following the experimental setup used in creating saliency datasets, we denote the decoder fixation queries as $\mathcal{F}_{q} = {F_{q0}, ...., F_{qN}}$. These queries simulate the viewer's attention, i.e.\ where they attend to the image features, and saccade to a spatial position that maximizes the attention task.  Furthermore, we do not want the queries to predict the same ground truth fixations, thus, the loss assigns a unique matching using the Hungarian algorithm between the prediction and ground truth fixations, and then minimizes the $l_1$ distance for the respective matched positions.

 Denote $f_{i} = (x_{i}, y_{i})$ as a sample from the ground truth set, and the corresponding prediction as $\bar f_{i} = (\bar x_{i},\bar y_{i})$. To find a bipartite matching between these two sets, as in \cite{carion2020end}, we search for a permutation of $N$ elements $\phi \in S_N$ with the lowest cost:
\begin{equation}
\hat{\phi} = \arg\min_{\phi \in S_N} \frac{1}{N} \sum_{i=1}^{N} \mathcal{L}_{\text{match}}(f_i, \hat{f}_{\phi(i)}),    
\end{equation}
where $\mathcal{L}_{\text{Match}}(f_i, \hat{f}_{\phi(i)})$ is a pairwise matching cost between the ground truth $f_i$ and a prediction with index $\phi(i)$. The final loss is: 
\begin{equation}
  \mathcal{L} = \sum_{i=1}^{N} \left| \left| f_i - \hat{f}_{\hat{\sigma}(i)} \right| \right|_{1}  + \alpha \mathcal{L}_{\text{NSS}} (\mathcal{M}_{p}, \mathcal{M}_{gt}),
  \label{loss}
\end{equation}
where $ \mathcal{M}_{gt} \in \{0,1\}^{H \times W}$ is the ground truth fixation, and  the predicted fixation map $ \mathcal{M}_{p}$ is obtained using:
\[\mathcal{M}_{p_{ij} }= \begin{cases}
1 & \quad \text{if location}~(i, j)~\text{is a fixation}\\
0 & \quad \text{otherwise},
\end{cases}\]
The normalized scanpath saliency loss (NSS) is defined as follows:
\begin{equation}
\mathcal{L}_{\text{NSS}} (\mathcal{M}_{p}, \mathcal{M}_{gt}) = \frac{1}{S} \sum_{i} \hat{\mathcal{M}}_{p_{i}}  \times \mathcal{M}_{gt_{i}},    
\end{equation}
where \(S = \sum_i \mathcal{M}_{gt_{i}}\) and \(\hat{\mathcal{M}}_{p} = \frac{{\mathcal{M}_{p} - \mu(\mathcal{M}_{p})}}{{\sigma(\mathcal{M}_{p})}}\), and \(i\) refers to the \(i\)-th pixel, and \(S\) represents the total count of fixated pixels. A score of $0$ indicates chance, while a positive NSS value indicates agreement between the maps beyond chance. Thus, we aim at minimizing the negative value of $\mathcal{L}_{\text{NSS}}$, by setting $\alpha$ to a negative value (-0.2).

\section{Experimental setting}

 \textbf{Training.} To evaluate the proposed framework, we train SalTR on the 10k/5k train/validation splits of the image saliency dataset Salicon~\cite{jiang2015Salicon}.

\textbf{Sampling the target fixations.} The ground truth fixation map $\mathcal{M}_{gt}$ contains a large number of fixations points (i.e., up to 500) gathered across a batch of subjects (16 viewers per image for the Salicon dataset). Hence, it's necessary to select $N$ points to match the number of fixation queries in the transformer decoder. Moreover, $N$ should be highly smaller than 500, for computational efficiency. For simplicity, we adopt a uniform sampling of $N$ samples out of $S$ points of $\mathcal{M}_{gt}$ to obtain $f_{i} = (x_{i}, y_{i})$, where $i \in \{1, \ldots, N\}$. This guarantees a diverse set of samples from the different viewers.

\textbf{Accelerating the training.} We observed that SalTR is difficult to optimize, and suffers from slow convergence, i.e., more than 100 epochs are needed to obtain comparable performance to baselines. Following the object detection literature \cite{dai2021up, sun2021rethinking, wang2021anchor, wang2021anchor}, several hypotheses can be proposed to account for this. First, the attention weights are uniformly assigned to all pixels in the feature maps at initialization, hence attending to meaningless locations that do not contribute to the feature propagation mechanism. Second, the discrete bipartite matching is unstable under stochastic optimization, as the same query is matched with different objects across epochs. Lastly, the decoder cross attention is under optimized in the early training, resulting in noisy contextual information for the queries. Inspired by the concept of deformable convolution \cite{dai2017deformable}, the approach of \cite{zhu2020deformable} is to add a translation term into the formula of the transformer attention, allowing a sparse spatial sampling by attending to a smaller set of locations (reference points). Consequently, this gating mechanism approximates the full self-attention via the locality inductive bias excluding potential long-term dependencies from the calculation. We adapt the deformable attention mechanism to our settings, termed Deformable SalTR.

\textbf{Evaluation setting.} We compare against the SoTA methods listed in \cite{wang2018revisiting} and add newer models with available implementations \cite{kummererSaliencyBenchmarkingMade2018}. Moreover, we test on the MIT300 benchmark \cite{Judd_2012}, which is more challenging than the Salicon test set. As suggested in \cite{kummererSaliencyBenchmarkingMade2018, salMetrics_Bylinskii}, we use the following evaluation metrics:  Similarity Metric (SIM), shuffled AUC (s-AUC), Linear Correlation Coefficient (CC), Normalized Scanpath Saliency (NSS), and the Kullback Leibler Divergence (KLD)~\cite{salMetrics_Bylinskii}. We adopt the more recent metrics formulation from \cite{kummererSaliencyBenchmarkingMade2018}.

\textbf{Technical details.} SalTR is implemented in PyTorch~\cite{pytorch} and trained using a single NVidia RTX3090 24GB GPU. We consider two configurations: SMALL with $3$ transformer encoder-decoder layers and BASE with $6$ layers, both with a ResNet-50 backbone. The number of fixation queries is set to 100, hence, 100 fixation points aresampled from the ground truth fixation map. SalTR variants are trained for $100$ epochs using the AdamW~\cite{adamw} optimizer, whereas the Deformable SalTR is trained for 40 epochs. We employ a warmup of $10$ epochs and a Cosine learning rate scheduler with maximum $lr$ set to $10^{-3}$.

\subsection{Results}

\begin{figure*}
\makebox[\linewidth]{
    \centering
    \includegraphics[width=0.99\linewidth]{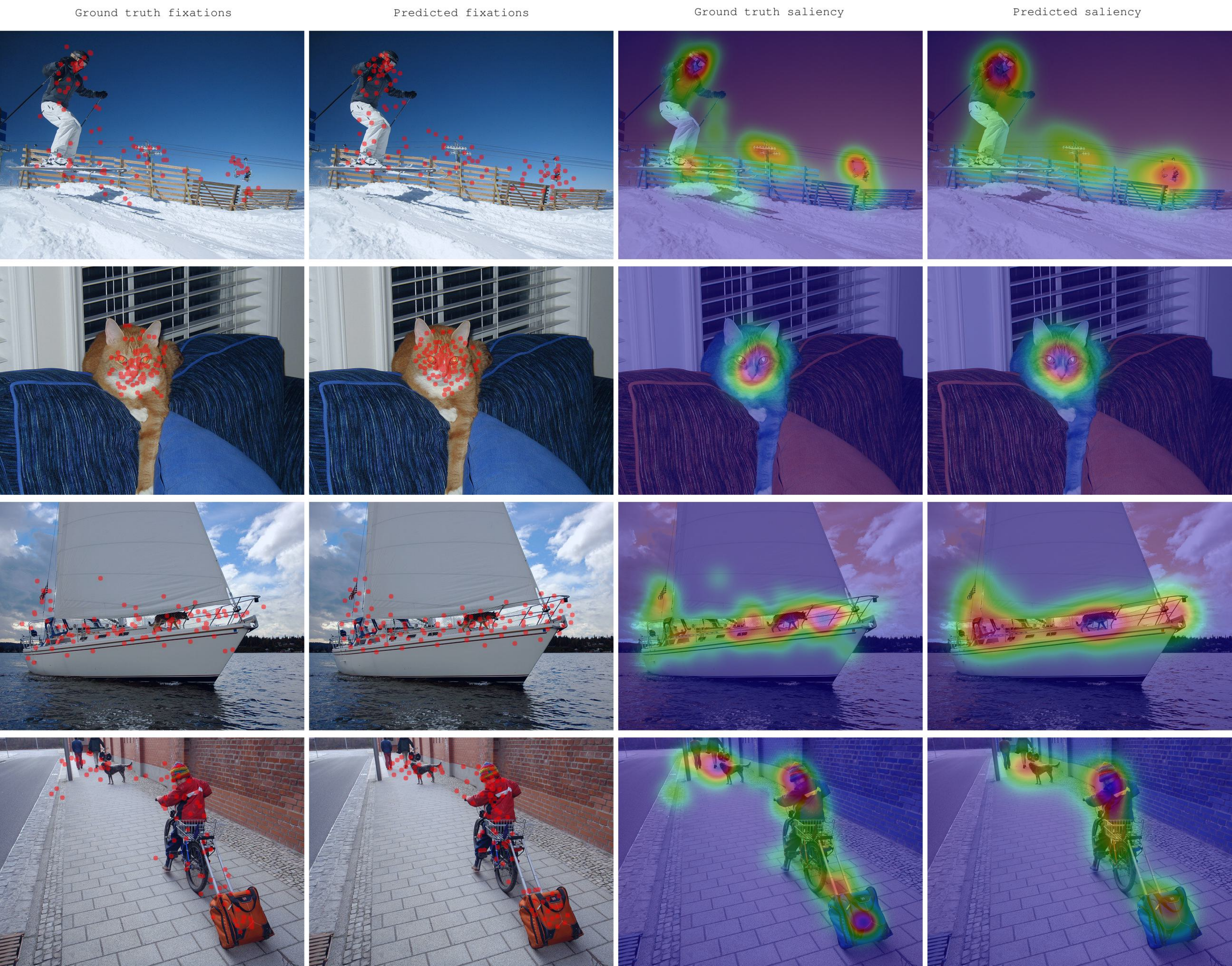}}
    \caption{Qualitative results of our model on sample images from SALICON. It can be observed that the proposed approach is able to handle various challenging scenes well and produces consistent fixation/saliency maps.}
    \label{results_fig}
\end{figure*}

\textbf{State-of-the-art comparison.} Here we compare the proposed approach to SoTA image saliency models on both the Salicon and MIT300 validation sets. \reftab{results1} shows the performance comparison in terms of the five metrics for the respective validation sets. We observe that our method performs favorably against existing approaches. 

\textbf{Salicon.} SalTR-Base achieves similar scores to UNISAL, whereas the Small variant showed a slightly worse performance highlighting the importance of the transformer decoder depth to optimize appropriately for the unstable bipartite matching. Using the same amount of compute, Deformable SalTR-Small matches the UNISAL performance due to the effective gated attention mechanism allowing for a smoother optimization. Conversely, Deformable SalTR-Base exhibits SoTA scores across the five metrics on par with the best model on these test examples (i.e. EML-NET~\cite{jia2020eml})

\textbf{MIT300.} The model was not trained on this dataset, making this an out-of-distribution test. The proposed models produces a reasonable improvement in accuracy compared to other models, except UNISAL and DeepGaze, which were trained on this specific dataset.

\reffig{results_fig} illustrates the predictions on a sample of the Salicon validation set. It can be seen that the fixation maps generated by our model (SalTR-Base) correlate well with the ground truth fixation maps in terms of fixation distribution. Also, we smooth our fixation maps with a Gaussian filter to obtain the continuous saliency map, giving the approach the flexibility to set the standard deviation parameters to fit the downstream application. Also, the effectiveness of the model in saccading to the main objects in the scene can be observed (see Figure 1,2 in supplementary). This demonstrates the effectiveness of the fixation queries in playing the role of the human subjects.

\begin{figure*}
\makebox[\linewidth]{
    \centering
    \includegraphics[width=0.99\linewidth]{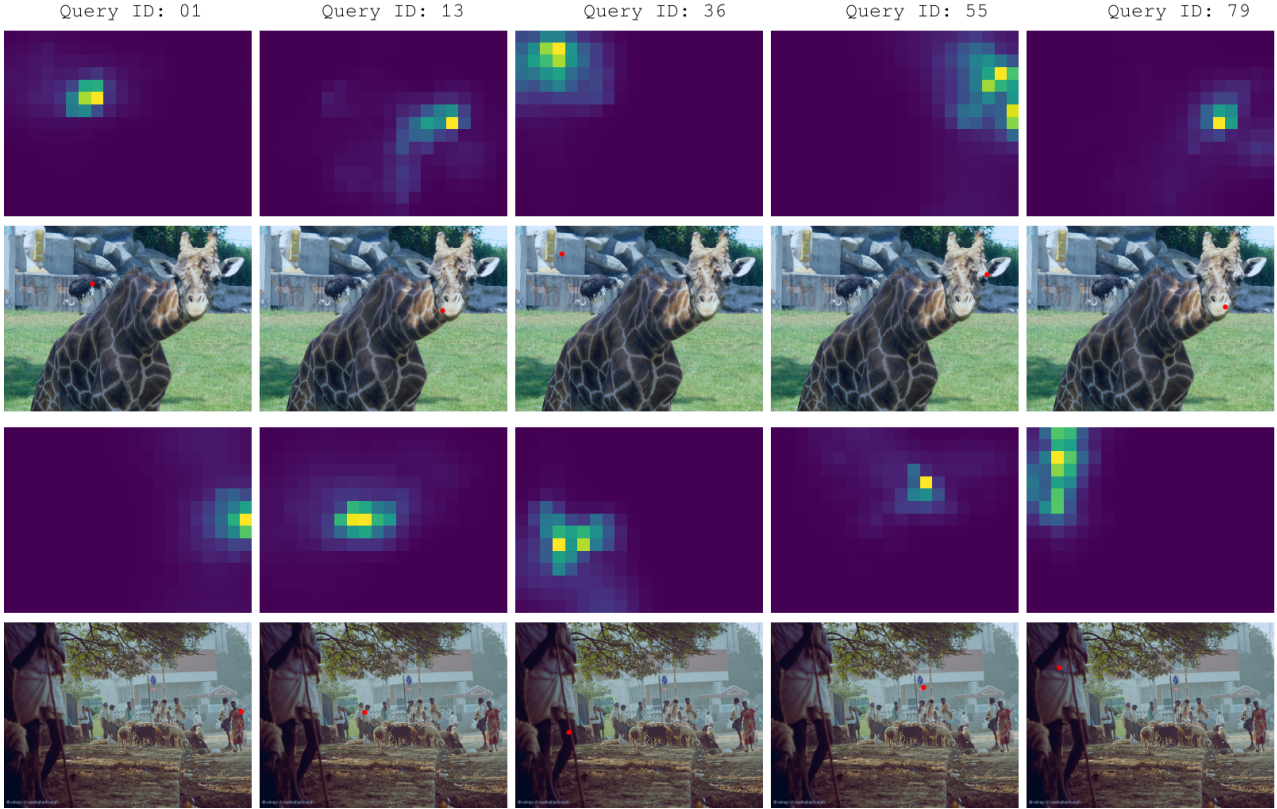}}
    \caption{Decoder self-attention for a set of fixation queries. It can be seen that queries attend to spatial locations consistent with their predicted fixation points. Predictions are made with SalTR-Base on a validation set of images.}
    \label{query_fig}
\end{figure*}

Furthermore, we visualize the decoder self-attentions maps for a set of a randomly selected queries with their respective fixation predictions in \reffig{query_fig}. We notice that each query's attention is highly local, where it attends to granular details of the image (such as: the Giraffe's mouth for query 13, and the person's leg for query 36). We believe that given well structured keys and values from the encoder latent representations, the queries focus on the refined granularities to predict the fixation points. This design is intuitive and is the closest to the experimental setting in obtaining the saliency datasets.

 \textbf{Saliency for low-level features.} SoTA saliency models capture high-level features such as cars, humans, etc. However, these kinds of approaches may fail to adequately capture a number of other crucial features that describe aspects of human visual attention that have been extensively investigated in psychology and neuroscience. Visual search, is one of the most prominent processes shaping human attention~\cite{treisman1980feature, kotseruba2020saliency}. This is where a subject's brain parallel processes regions that differ significantly in one feature dimension i.e.\ color, intensity, orientation. These correspond to low-level features, which operate as the basic mechanisms of the human visual system. We conducted evaluations of the performance of UNISAL, and our SalTR on samples of low-level attention using images from a recently proposed dataset~\cite{kotseruba2020saliency}. The aim is to understand the main differences on how saliency exploration is performed when the self-attention mechanism promotes global connectivity between the image patches. See Section A.3 in the supplementary materials for more details.

As shown in Figure.2 in supplementary, UNISAL produces high-quality saliency maps consistent with the ground truth maps for natural images from Salicon.  High-level features such as: human faces, bus, monument, etc; are dominant in these images. The human visual system combines the bottom-up with top-down features to solve the attention task. This behaviour might not be reflected in the fixation/saliency datasets. Hence, end-to-end deep learning based models might learn a good saliency mapping, but actually violate the subtleties of its true definition. Early computational approaches for the visual human system e.g.~\cite{navalpakkam2006integrated,kootstra2008paying,murray2011saliency,bruce2006saliency,garcia2009decorrelation,seo2009static,goferman2012context,gao2005discriminant} were mostly cognitive based models relying on computing multiple visual features such as color, edge, and orientation at multiple spatial scales to produce a saliency map. Moreover, could the fixation queries in SalTR bridge this gap, by reasoning over the recurring features and drawing dependencies/similarities?

In fact, UNISAL fails to respond to simple features. For example, considering colour (Figure.3 in supplementary), UNISAL~\cite{droste2020unified} did not capture the penguin as the most salient object, whereas the SalTR succeeded in doing so, as this pattern is solved with the global nature of the self-attention mechanism. This suggests that SalTR is better at incorporating characteristics of the visual system  as important priors induced by the self-attention mechanism and the Hungarian matching. Clearly however, as shown in Figure.4 in supplementary, SalTR severely fails when given synthetic images, the model does not respond well to low-level features from the O3 dataset, and nearly produces random fixations around the center.

 \begin{table*}
 \caption{Comparative performance study on Salicon and MIT300.  }

 \begin{tabular}{ll | cllll | cllll}
        \toprule
      \multicolumn{2}{c}{Models} &\multicolumn{5}{c}{\textbf{Salicon}} &
      \multicolumn{5}{c}{\textbf{MIT300}}\\
        
      &&SIM &s-AUC& CC& NSS& KLD& SIM& s-AUC& CC& NSS& KLD\\

 \midrule
 \multirow{2}{*}{}

&ITTI ~\cite{itti1998model} & 0.37 &  0.61 & 0.20 & -- & --  & 0.46 &  0.13 & 0.44 & 1.11 & 0.95  \\

&GBVS~\cite{harel2006graph} & 0.44 &  0.63 & 0.42 & -- & --  & 0.48 &  0.62 & 0.47 & 1.24 & 0.88 \\

&Salicon ~\cite{jiang2015Salicon} & -- &  -- & -- & -- & --  & 0.51 &  0.73 & 0.56 & 1.70 & 0.78 \\

&CASNet ~\cite{fan2018emotional} &  &  -- & -- & -- & --  & 0.58 &  0.73 & 0.70 & 1.98 & 0.58 \\

&EML-NET~\cite{jia2020eml} & 0.79 &  0.74 & 0.89 & 2.05 & 0.52  & 0.74 &  0.67 & 0.78 & 2.48 & 0.84 \\

&MSI-Net ~\cite{kroner2020contextual} & 0.80 &  0.74 & 0.90 & 2.01 & & 0.67 &  0.74 & 0.77 & 2.30 & 0.42\\

&TranSalNet ~\cite{lou2022transalnet} & -- &  -- & -- & -- & -- &  0.68 & 0.74& 0.80 & 2.41 &1.01 \\

 &SalGAN~~\cite{pan2017salgan}&  &0.75 & 0.76& 2.47& - &  0.63& 0.73& 0.67 & 1.86& 0.75 \\

 &UNISAL~~\cite{droste2020unified} &0.77& 0.73& 0.87& 1.95& - & 0.67& 0.78& 0.78& 2.36& 0.41\\

 &DeepGaze~~\cite{kummerer2017understanding} & -- &  -- & -- & -- & -- & 0.66& 0.77& 0.77& 2.33& 0.42\\

\midrule

 &SalTR-Small & 0.75 & 0.71  & 0.84 & 1.79 & 0.98 & 0.61 & 0.70& 0.70& 1.98& 0.59 \\
 &SalTR-Base & 0.78 &  0.75 & 0.87 & 1.97 & 0.74 & 0.65& 0.74& 0.75& 2.22& 0.46 \\
 
\midrule
&Deformable SalTR-Small & 0.77 &  0.73 & 0.86 & 1.88 & 0.82 & 0.64& 0.75& 0.76& 2.09& 0.49 \\
&Deformable SalTR-Base & 0.79 &  0.77 & 0.89 & 2.12 & 0.62 &  0.69& 0.79& 0.80& 2.45& 0.36 \\

 \bottomrule
 \end{tabular}
 \label{results1}
 \end{table*}

\subsection{Ablation study}

\textbf{Losses importance.} The bipartite matching loss appears intuitive to avoid fixation queries collapse. We validate this hypothesis by eliminating the Hungarian matching loss, and only use the final loss in \refeq{loss} without any assignments. As expected, the model learns the saliency dataset center bias. In other words. the fixation queries tend to all focus on the most salient features of the input image, while ignoring the rest. This is an artifact of the cross-attention optimization, where a shortcut over the features dominates learning more diverse and rich predictions.

\textbf{Object detection vs saliency prediction.} We attempted to initialize SalTR with DETR weights trained on COCO for object detection. We only train the MLP head on the saliency dataset. The aim is to measure the alignment between the two downstream tasks. The model was unable to achieve the baseline scores (i.e.\ KLD: 2.5, NSS: 0.9). We observed that the fixations were mostly over-estimated around the objects center (see Figure 3 in supplementary). We hypothesize that the salient regions in an image may correspond to the objects of interest within the image. Clearly, however, the most salient regions are not necessarily the objects in the image, but could rather be other features or patterns that catch the viewer's attention~\cite{borji2012state}. This potentially explains the failure of this setting.

\textbf{Varying the number of fixation queries.} We investigate the optimal number of target fixations for better  results. As shown in \reftab{results2}, we vary the number of queries $N$, hence the number of target fixations $S$. We observe that 100 is the optimal value and higher values result in a diminishing returns in the performance since higher number of queries make the optimization harder.

\textbf{The impact of the Gaussian smoothing.} \reftab{results_sigma} represents the impact of the standard deviation ($\sigma$) of a Gaussian filter applied to the fixation map for the SalTR-Base model. It can be observed that the $\sigma = 19.0$ generally offers the best performance across most metrics. Notably, the model achieves the highest SIM, CC, and NSS scores at $\sigma = 19.0$. Furthermore, we can observe that  the KLD is the metric that is highly affected by the smoothing parameters, whereas the CC is more or less the same after the 19.0 value.

\begin{table}
 \caption{Impact of the number of fixation queries. Performance comparison when it is varied. 100 is the optimal number of queries for the saliency task. SalTR-Base is showed.}

 \adjustbox{width=\columnwidth}{
 \begin{tabular}{ll | cllll}
        \toprule
      \multicolumn{2}{c}{SalTR} &\multicolumn{5}{c}{\textbf{Salicon}} \\
        
      &&SIM &s-AUC& CC& NSS& KLD\\

 \midrule

 \multirow{2}{*}{Number of queries}

 &50& 0.75 & 0.70  & 0.82 & 1.55 & 1.04   \\

 &\textbf{100}& 0.78 &  0.75 & 0.87 & 1.97 & 0.74\\

 &150& 0.79 &  0.74 & 0.88 & 1.92 & 0.71 \\
 
 &200& 0.78 &  0.74 & 0.86 & 1.94 & 0.78 \\ 

 \bottomrule

 \end{tabular}}
 \label{results2}
 \end{table}
 
\textbf{All vs single subject.} To the best of our knowledge, SalTR is the first approach for learning saliency prediction from fixations only. However, it still does not replicate the the experimental setting fully, as the fixation queries predict a single locations, whereas human subjects may attend to multiple locations. We attempted to sample a single subject as targets so all the queries simulates a single human with multiple predicted fixations. This design resulted in visually appealing saliency maps (see Figure 6 in supplementary), but the scores did not match the baselines because the predictions were mostly sparse. This potentially highlights an issue with the metrics, that require an over-estimated saliency map to match the dense ground truth map aggregated over a batch of subjects.

\subsection{SalTR for scanpath prediction}
Scanpath prediction refers to the process of estimating the timely trajectory of fixation points for a human subject when viewing a visual stimulus. The SalTR design allows for manipulating the Transformer decoder mask. Indeed, we used the full mask for parallel decoding previously; clearly, however, we can use a causal mask, and add an end-of-sequence (EOS) (i.e.\ position $(0, 0)$) to the target fixations, for the auto-regressive decoding. By doing so, SalTR can naturally handle ordered fixation predictions (i.e.\ scanpaths). The prediction head outputs the fixations points and their duration in seconds. We train SalTR-Base  using the 10k Salicon images; for each image, we select all the human subject's scanpaths, each serving as its own separate training example. The model should converge on the correct distribution over scanpaths given the image. We finetune the transformer decoder only for 20 epochs.

\textbf{Results.} The Multi-Match (MM) \cite{dewhurst2012depends} measure is used as the main metric for ranking scanpath prediction models. Our approach obtains an MM score of 0.93 on the 5k Salicon examples (PathGAN~\cite{assens2018pathgan}: 0.96). This is achieved by averaging the individual results against all human viewer scanpaths for a single image. Figure 4 in the supplementary material shows the quantitative results on sample images from Salicon. Our model demonstrates a clear and consistent tracking of the ground truth, indicating its ability to accurately capture human visual attention.

\begin{table}
 \caption{Impact of the number standard deviation. Performance comparison when the Gaussian smoothing parameter is varied. SalTR-Base is showed.}

 \adjustbox{width=\columnwidth}{
 \begin{tabular}{l | cllll}
        \toprule
      &\multicolumn{5}{c}{\textbf{Salicon}} \\
        
      &SIM &s-AUC& CC& NSS& KLD\\

 \midrule

 5.0& 0.42 & 0.50  & 0.54 & 1.33 & 1.63   \\

 10.0& 066 &  0.70 & 0.77 & 1.77 & 0.99 \\

\textbf{19.0}& 0.78 &  0.75 & 0.87 & 1.97 & 0.74\\

 30.0 & 0.76 &  0.72 & 0.87 & 1.78 & 1.16\\ 

 \bottomrule

 \end{tabular}}
 \label{results_sigma}
 \end{table}

\section{Conclusion}

 We present a novel approach for saliency prediction in images, named Saliency TRansformer (SalTR), which leverages parallel decoding in transformers to learn saliency solely from fixation maps. Unlike existing models that rely on continuous saliency maps, our approach directly predicts fixations by treating saliency prediction as a set prediction problem. We conducted experiments on the Salicon and MIT300 benchmarks and achieved remarkable performance compared to SoTA methods. Our approach not only replicates the data collection pipeline used in generating saliency datasets but also eliminates the need for continuous saliency annotations. Furthermore, we extended our approach to the scanpaths prediction problem and demonstrated its effectiveness. Overall, our approach offers a promising direction for saliency prediction, focusing on the discrete fixation maps and directly predicting fixation points. It opens up possibilities for application-guided saliency prediction, as per the flexibility offered in our design.

{\small
\bibliographystyle{ieee_fullname}
\bibliography{egbib}
}

\clearpage
\appendix

\setcounter{table}{0}
\setcounter{figure}{0}
\renewcommand{\thetable}{A.\arabic{table}}
\renewcommand{\thefigure}{A.\arabic{figure}}

\section*{\LARGE Appendices}

In this supplementary, we present additional analysis related to our proposed method SalTR. 

\section{Additional results}
\begin{figure*}
\makebox[\linewidth]{
    \centering
    \includegraphics[width=1.00\linewidth]{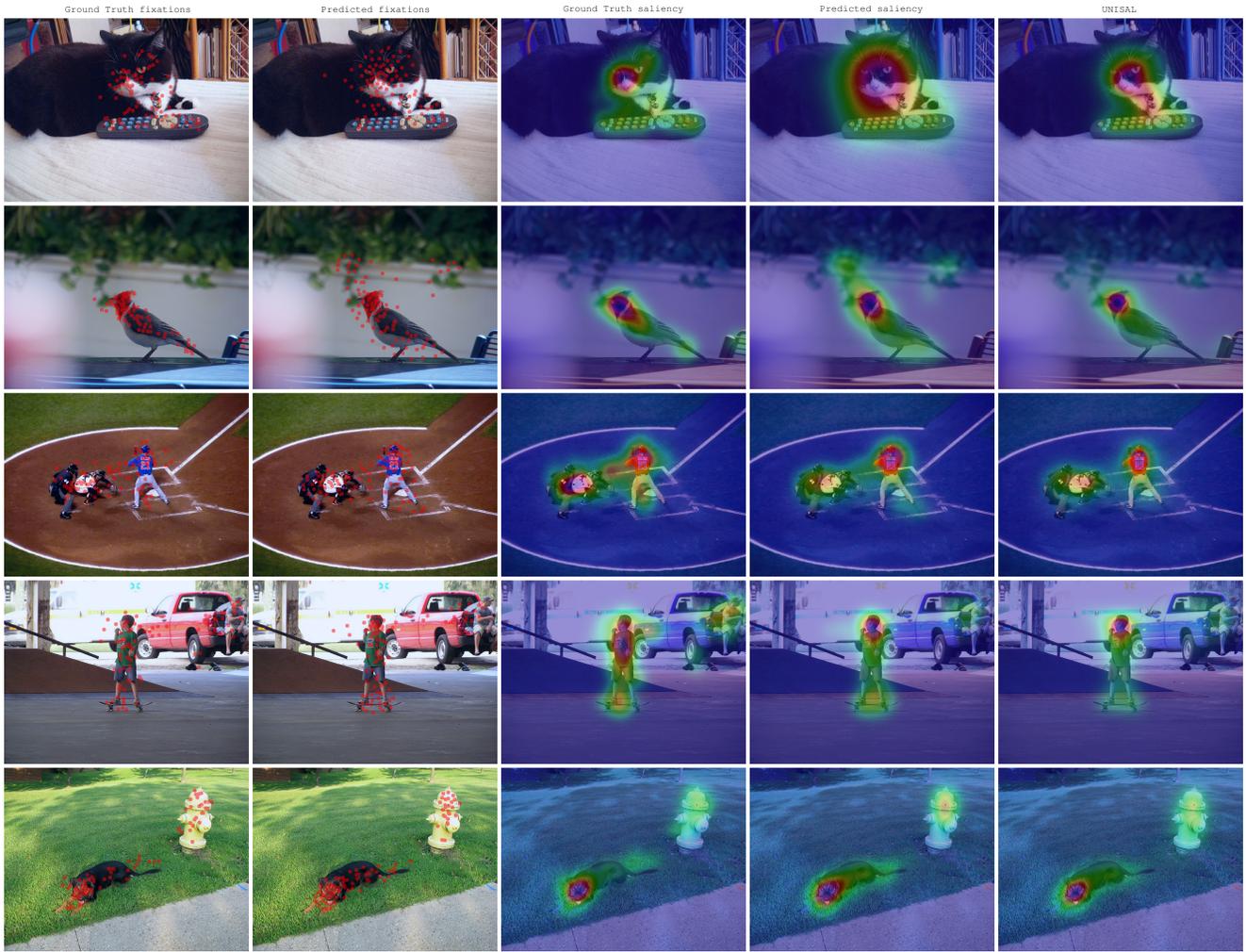}}
    \caption{Qualitative results on the Salicon validation images against Unisal.}
    \label{fig1}
\end{figure*}

Extra qualitative results on saliency prediction are shown in \reffig{fig1}, it can be seen that SalTR produces high quality fixation and saliency maps compared to the Ground Truth ones. Furthermore, Unisal \cite{droste2020unified} predicts good saliency maps as well, hence, both models demonstrate their robustness across a variety of image complexities. 

\begin{figure*}
\makebox[\linewidth]{
    \centering
    \includegraphics[width=1.00\linewidth]{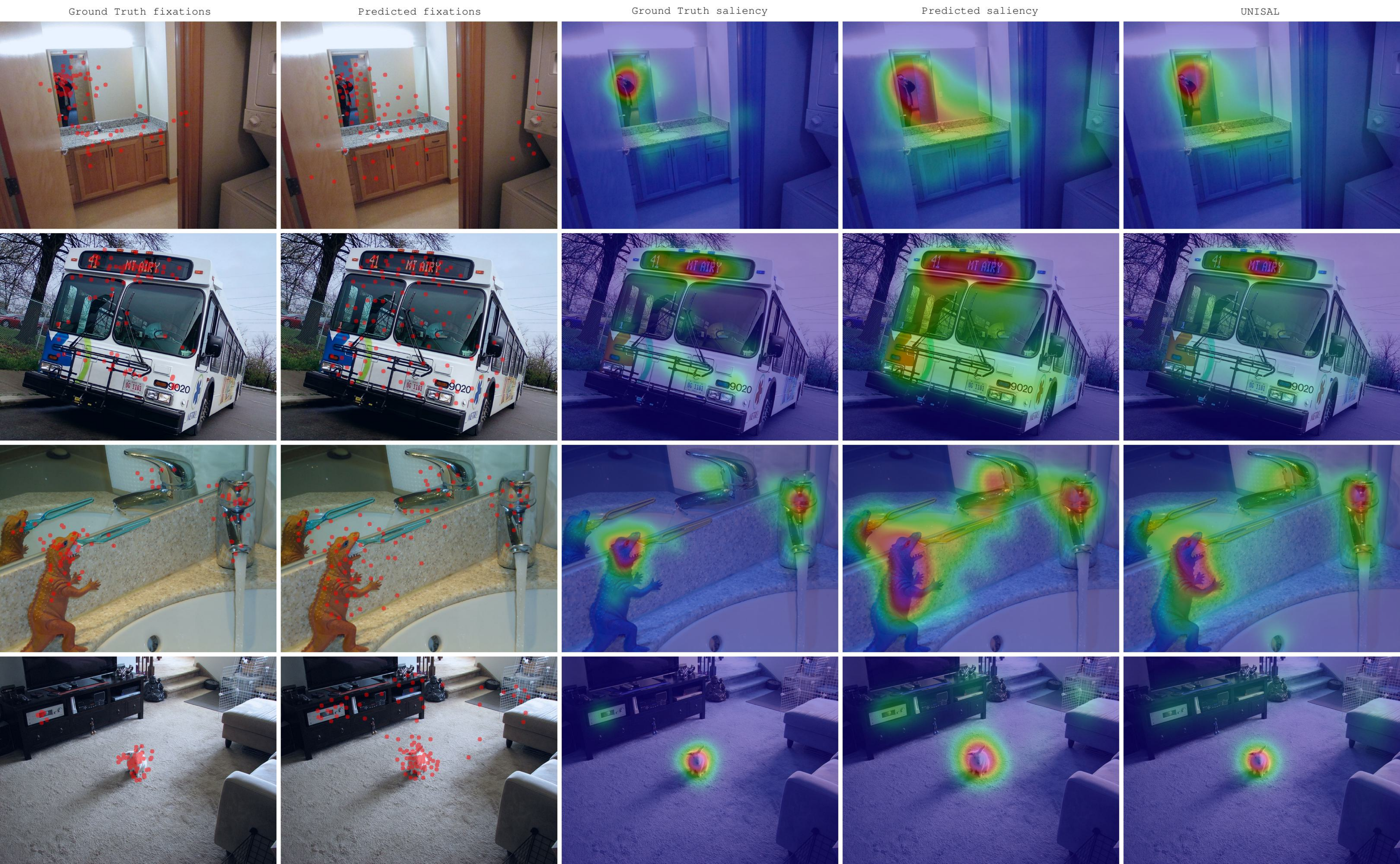}}
    \caption{Images from the Salicon validation set where SalTR obtains high KLD scores.}
    \label{fig2}
\end{figure*}

\textbf{Failure cases.} In an effort to delve deeper into the SalTR capabilities, we have identified certain anomalies within the Salicon validation set, specifically images marked by the highest Kullback-Leibler Divergence (KLD) value of 1.51 and the lowest Normalized Scanpath Saliency (NSS) score of 1.09. As visualized in \reffig{fig2}, we have so far pinpointed two areas where the model did not perform as expected, which we refer to as \textit{failure modes}. In the top image, the model fails to sharpen the prediction around the human, that can be considered as a small object. The lack of hierarchical decoding in our framework may be the reason behind this failure, it is important to consider decoding multiple resolutions of the latent representations to allow the SalTR to adapt to multiple scale fixation regions. A second failure we observe in the last image, where the Ground Truth fixations are all concentrated around the rabbit in the center. The Hungarian matching forces distinct predictions, thus, it favors more spread out predictions. This behaviour is overcomed by the decoder when the input image does not account any other candidate salient regions. Clearly however, when the decoder picks other informative areas in this scenario, it tends to uniformly assign fixations.

\textbf{No bipartite matching.} \reffig{main_fig} illustrates the results of the SalTR model trained without the incorporation of the Hungarian matching loss. As can be observed, the predictions lack diversity and are all centered around the same region, akin to a duplicated query trained for 100 epochs. The lack of variety in the predicted regions suggests that the model is picking a specific feature or set of features in the training data. In the absence of the Hungarian matching loss, which usually serves to optimize assignment between predictions and ground truth, the SalTR model seems to struggle with providing unique, diversified predictions. The model appears to 'latch on' to a particular set of features or patterns, consequently producing a narrowed range of output. This behavior strongly suggests the crucial role of the Hungarian matching loss in facilitating SalTR's ability to make diverse predictions. Without it, the model's ability to generalize well across varying input data seems to be significantly hindered. The training duration of 100 epochs, in this case, does not appear to alleviate the observed issue.

\textbf{Low-level features.} We evaluate our model's performance using images from both the P3 and O3 datasets \cite{kotseruba2020saliency}. The model's performance on real images, while not perfect, is quite commendable, as illustrated in \reffig{low}. It successfully identifies the feature in question and assigns a higher saliency density to it. Concurrently, it maintains attention to other regions in the image, demonstrating an acceptable level of distribution in its focus. However, defining an ideal saliency map for such images is an intricate endeavor. It is generally agreed that humans tend to initially focus on the most visually prominent or 'bottom-up' features within the first moments of observation. Following this, exploratory eye movements typically begin, covering other areas of interest in the image. Capturing this dual-phase attention process is a challenging task that current models are yet to master comprehensively.

On the other hand, our model's performance on synthetic images, as displayed in \reffig{low_fake}, is noticeably weaker. The model appears to struggle with understanding and interpreting the abstract features typically presented in synthetic imagery. This shortcoming is possibly attributed to the learning process. Specifically, the Salicon training set, which our model was trained on, does not include any synthetic images. The model's ability to generalize from the real images in the training set to synthetic images in the test set seems to be a significant hurdle.

In light of these observations, future efforts might be geared towards including a more diverse range of image types in the training set, particularly synthetic images. Furthermore, more research is needed to improve the model's ability to mimic the human attention process more accurately, especially concerning the transition from bottom-up to exploratory attention.

\begin{figure*}
  \makebox[\linewidth]{
    \centering
    \resizebox{\linewidth}{1.0\textheight}{
      \includegraphics{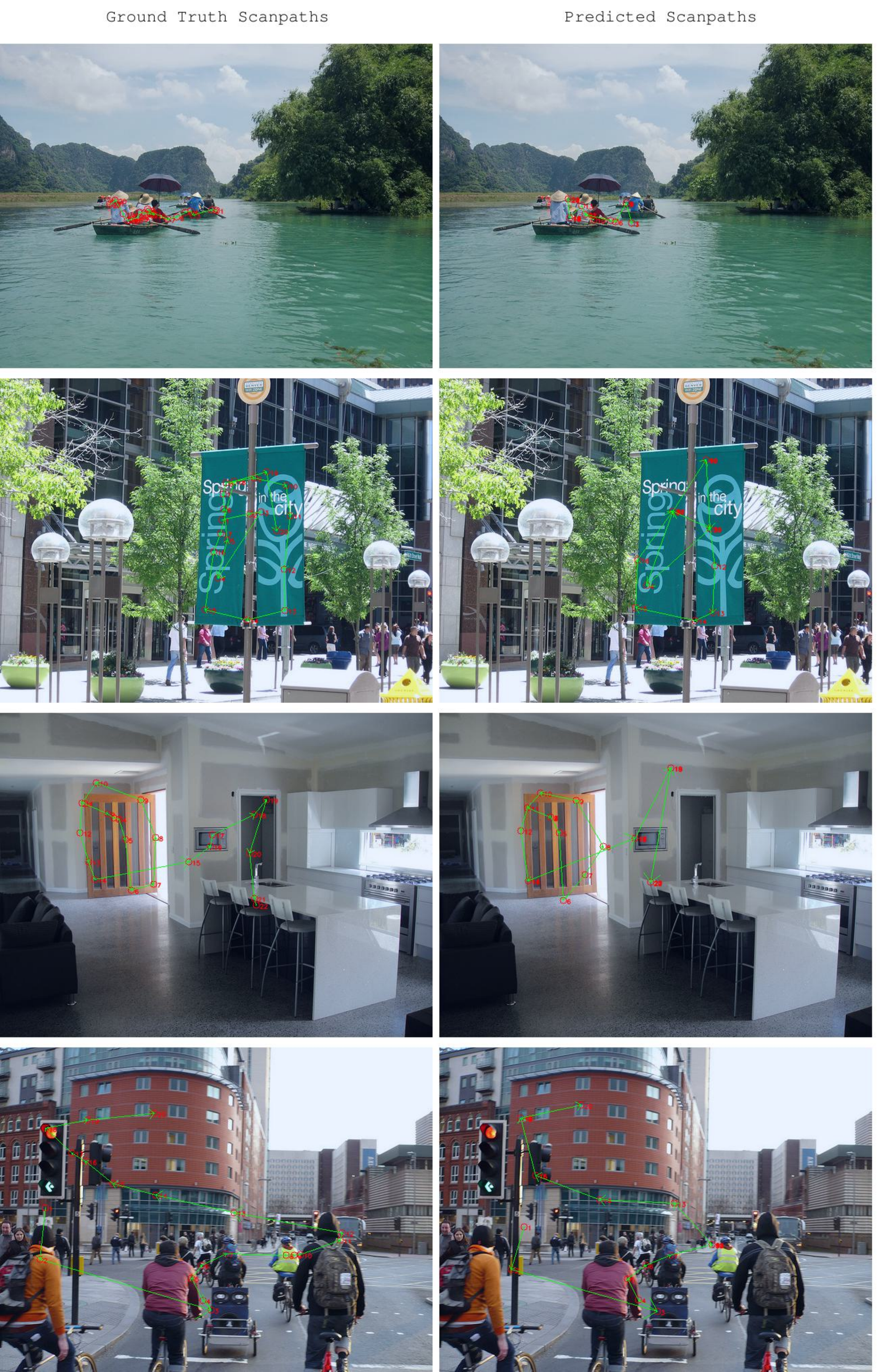}
    }
  }
  \caption{Scanpaths for a set of images from the Salicon validation set.}
  \label{results_fig_2}
\end{figure*}

\begin{figure*}
\makebox[\linewidth]{
    \centering
    \includegraphics[width=1.00\linewidth]{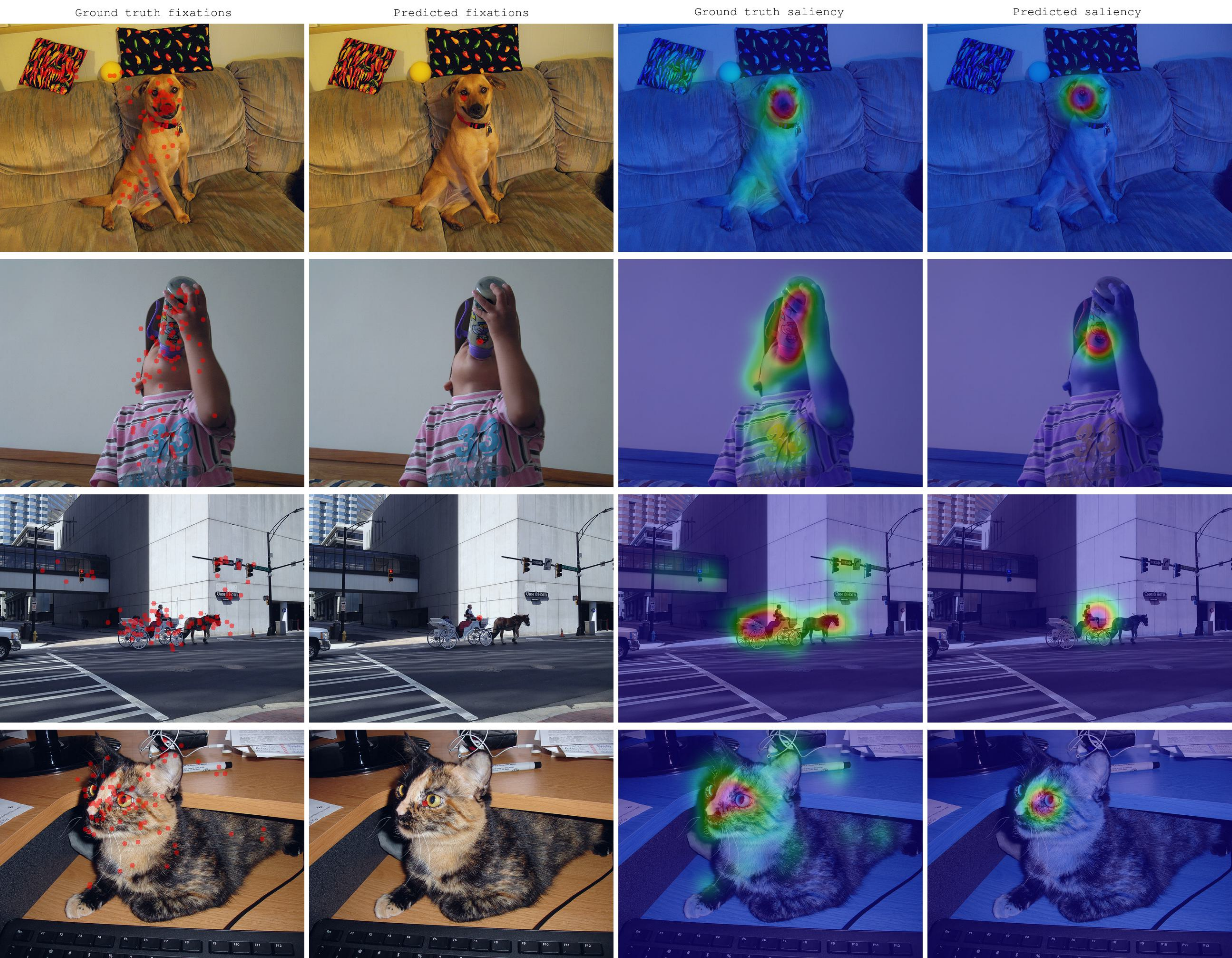}}
    \caption{Images from the Salicon validation set where the model was trained without any Hungarian matching loss.}
    \label{main_fig_sup}
\end{figure*}

\begin{figure*}
\makebox[\linewidth]{
    \centering
    \includegraphics[width=1.00\linewidth]{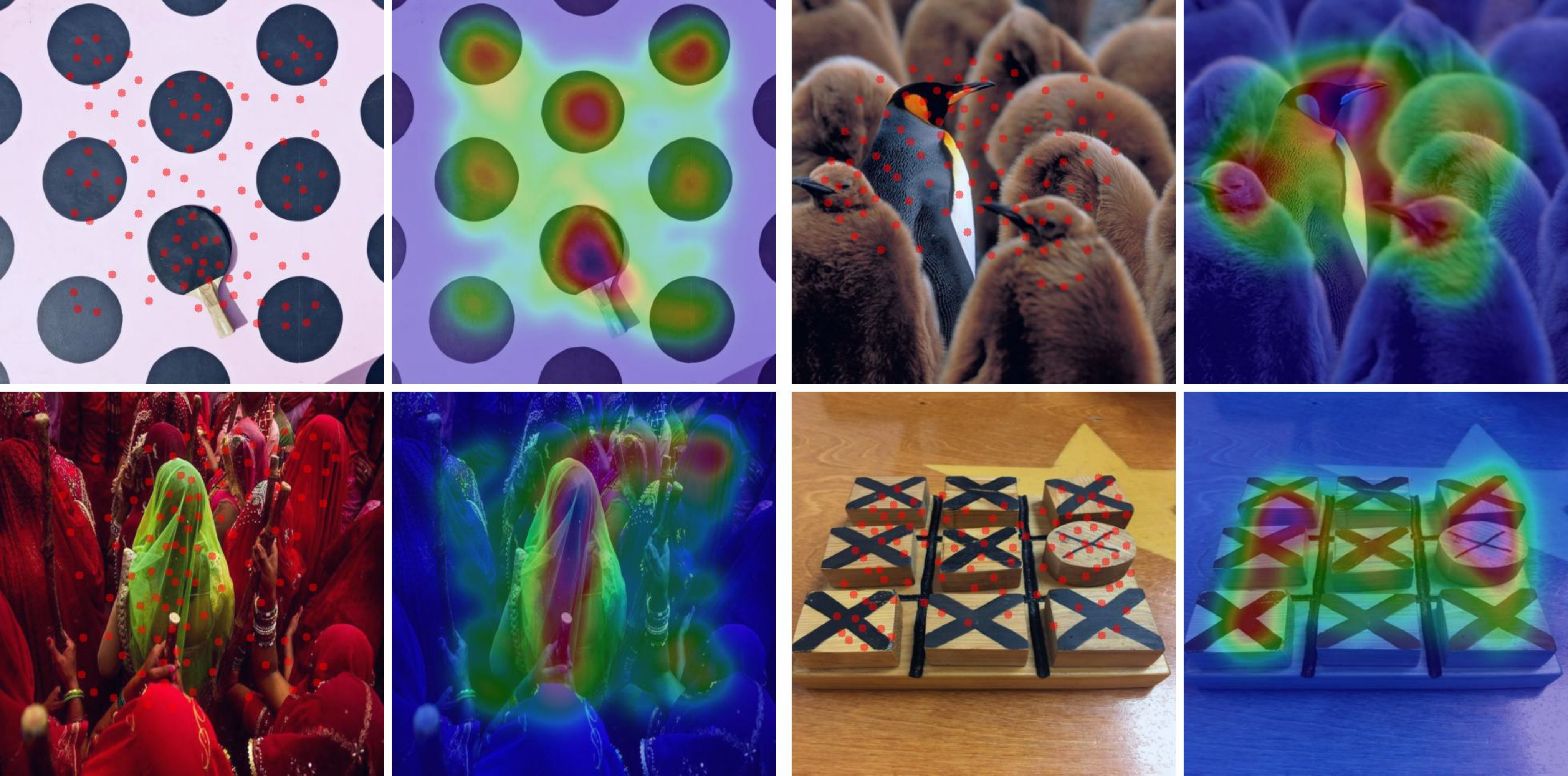}}
    \caption{Predictions on the low-level features P3 dataset.}
    \label{low}
\end{figure*}

\begin{figure*}
\makebox[\linewidth]{
    \centering
    \includegraphics[width=1.00\linewidth]{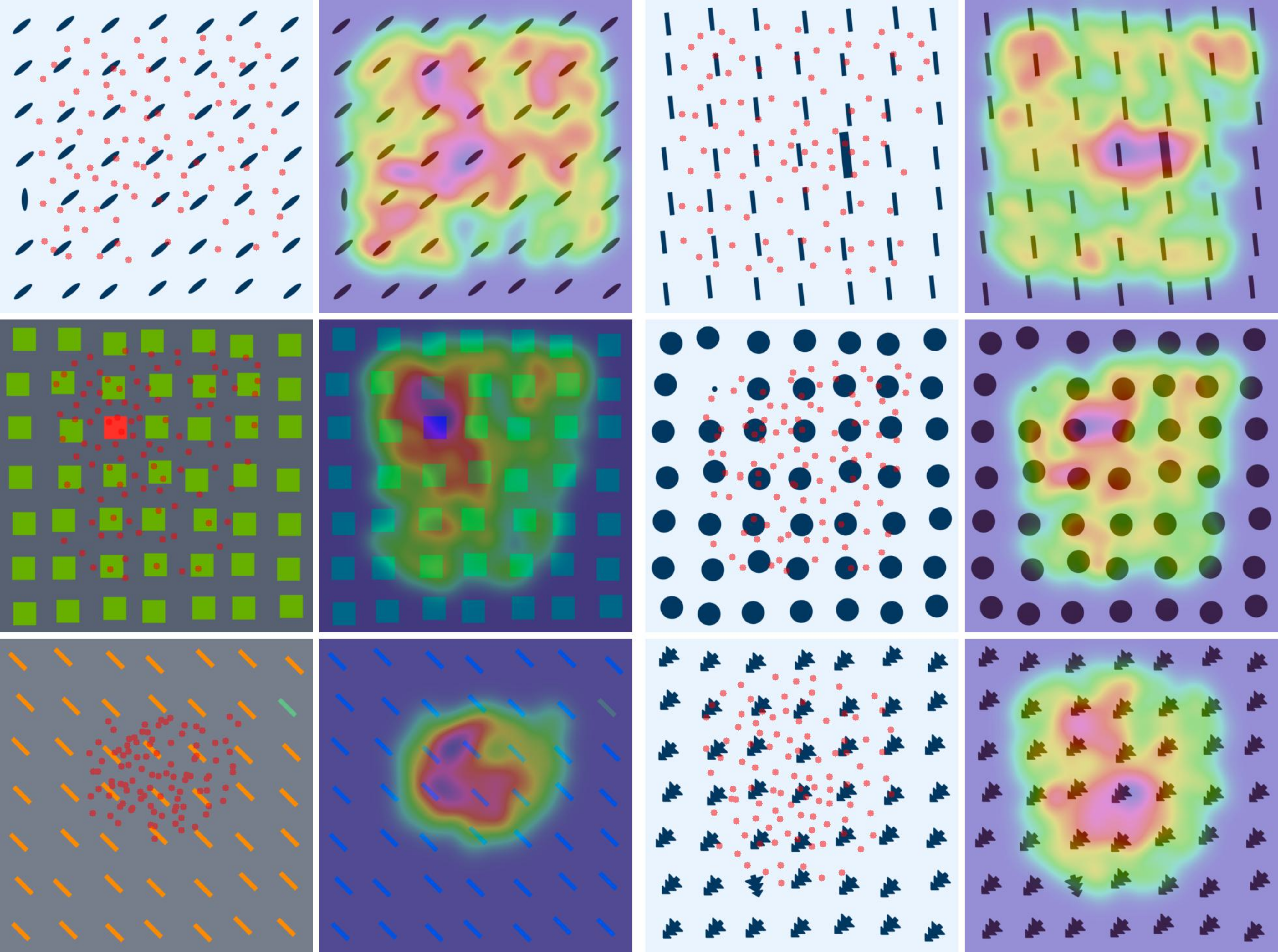}}
    \caption{Predictions on the low-level features O3 dataset.}
    \label{low_fake}
\end{figure*}

\end{document}